%% file: vts.tex
\documentclass{article}
\usepackage[margin=1.5in]{geometry}

\usepackage[utf8]{inputenc} 
\usepackage[T1]{fontenc}    
\usepackage{hyperref}       
\usepackage{url}            
\usepackage{booktabs}       
\usepackage{amsfonts}       
\usepackage{nicefrac}       
\usepackage{microtype}      

\usepackage{amsmath}
\usepackage{dsfont} 
\usepackage{graphicx}
\usepackage{caption}
\usepackage{subcaption}
\usepackage{float} 
\usepackage{booktabs} 
\usepackage{multirow} 
\usepackage[table]{xcolor}
\usepackage{flushend}
\usepackage{algorithm}
\usepackage{algorithmic}
\usepackage{tikz}
\usetikzlibrary{bayesnet} 

\usepackage[round,numbers,sort]{natbib}

\def \Real{{\mathbb R}} 
\def \Natural{{\mathbb N}} 
\newcommand{\eValue}[1]{\mathbb{E}\left\{ #1 \right\}}

\newcommand{\ie}{i.e., }

\newcommand{\eg}{e.g., }

\newcommand{\N}[1]{\mathcal{N}\left( #1\right)}

\newcommand{\Dir}[1]{{\rm Dir\left( #1\right)}}

\newcommand{\Cat}[1]{{\rm Cat}\left( #1\right)}

\newcommand{\NIG}[1]{{\rm NIG}\left( #1\right)}

\newcommand{\U}[1]{\mathcal{U}\left( #1\right)}

\newcommand{\deq}{:=} 

\newcommand{\argmax}{\mathop{\mathrm{argmax}}}

\newcommand{\tr}[1]{\mathrm{tr}\left\{ #1 \right\}} 

\title{Variational inference for the multi-armed contextual bandit}
	
\author{ I\~{n}igo Urteaga and Chris H.~Wiggins\\
	{\sf \{inigo.urteaga, chris.wiggins\}@columbia.edu} \\\\
	Department of	Applied Physics and Applied Mathematics\\
	Data Science Institute\\
	Columbia University\\
	New York City, NY 10027
}

\begin{document}
\maketitle

\begin{abstract}
In many biomedical, science, and engineering problems, one must sequentially decide which action to take next so as to maximize rewards. One general class of algorithms for optimizing interactions with the world, while simultaneously learning how the world operates, is the multi-armed bandit setting and, in particular, the contextual bandit case. In this setting, for each executed action, one observes rewards that are dependent on a given `context', available at each interaction with the world. The Thompson sampling algorithm has recently been shown to enjoy provable optimality properties for this set of problems, and to perform well in real-world settings. It facilitates  generative and interpretable modeling of the problem at hand. Nevertheless, the design and complexity of the model limit its application, since one must both sample from the distributions modeled and calculate their expected rewards. We here show how these limitations can be overcome using variational inference to approximate complex models, applying to the reinforcement learning case advances developed for the inference case in the machine learning community over the past two decades. We consider contextual multi-armed bandit applications where the true reward distribution is unknown and complex, which we approximate with a mixture model whose parameters are inferred via variational inference. We show how the proposed variational Thompson sampling approach is accurate in approximating the true distribution, and attains reduced regrets even with complex reward distributions. The proposed algorithm is valuable for practical scenarios where restrictive modeling assumptions are undesirable.
\end{abstract}

\section{Introduction}
\label{sec:introduction}

Reinforcement learning is an area of machine learning that studies optimizing interactions with the world while simultaneously learning how the world operates. The multi-armed bandit problem \cite{b-Sutton1998,j-Ghavamzadeh2015} is a natural abstraction for a wide variety of such real-world challenges that require learning while simultaneously maximizing rewards. The goal is to decide on a series of actions under uncertainty, where each action can depend on previous rewards, actions, and contexts, aiming at balancing exploration and exploitation.

The name ``bandit'' finds its origin in the playing strategy one must devise when facing a row of slot machines (\ie which arms to play). The setting is more formally referred to as the theory of sequential decision processes. Its foundations in the field of statistics began with the work by \citet{j-Thompson1933,j-Thompson1935} and continued with the contributions by \citet{j-Robbins1952}. Interest in sequential decision making has recently intensified in both academic and industrial communities. The publication of separate works by \citet{ic-Chapelle2011}, and \citet{j-Scott2015} have shown its impact in the online content management industry. This renaissance period of the multi-armed bandit problem has both a practical aspect \cite{ip-Li2010} and a theoretical one as well \cite{j-Scott2010,ip-Agrawal2012,ip-Maillard2011}.

Interestingly, most of these works have orbited around one of the oldest heuristics that address the exploration-exploitation tradeoff, \ie Thompson sampling. It has been empirically proven to perform satisfactorily, and to enjoy provable optimality properties, both for problems with and without context \cite{ip-Agrawal2012,ip-Agrawal2013,ip-Agrawal2013a,ic-Korda2013,j-Russo2014,j-Russo2016}.

In this work, we are interested in extending and improving Thompson sampling. In its standard form, it is applicable to restricted models of the world, as one needs to sample from the corresponding parameter posteriors and compute their expected rewards: see \cite{j-Scott2010} for details. The issue is that, for many problems of practical interest, one has partial (or no) knowledge about the ground truth, and the available models might be misspecified. 

We aim at extending Thompson sampling to allow for more complex and flexible reward distributions. We target a richer class of bandits than in the most recent literature, where the posterior is usually assumed to be from the exponential family of distributions \cite{ic-Korda2013}.

We model the convoluted relationship between the observed variables (rewards), and the unknown parameters governing the underlying process by mixture models, a large hypothesis space which for many components can accurately approximate any continuous reward distribution. The main challenge is how to learn such a mixture distribution within the contextual multi-armed bandit setting.

To that end, we leverage the advances developed for statistical inference in the last decades, and propose a variational approximation to the underlying true distribution of the environment with which one interacts. Variational inference is a principled framework, with roots in statistical physics and widely applied in the machine learning community \cite{b-Bishop2006}.

Approximation of Bayesian models by variational inference has already attracted interest within the reinforcement learning community, \eg to learn a probability distribution on the weights of a neural network \cite{ip-Blundell2015}. Thompson sampling has also been applied in the context of deep Q-networks, \eg \cite{ip-Lipton2018} and \cite{ic-Osband2016}. Nevertheless, our focus here is (a) not on Q-learning but on bandit problems, and (b), the variational inference is for a hierarchical Bayesian mixture model approximation to the true reward distribution. We show that variational inference allows for Thompson sampling to be applicable for complex reward models.

Our contribution is unique to the contextual multi-armed bandit setting in that (a) we approximate unknown bandit reward functions with Gaussian mixture models, and (b) we provide variational mean-field parameter updates for the distribution that minimizes its divergence (in the Kullback-Leibler sense) to the mixture model reward approximation.

The proposed method autonomously learns, in the contextual bandit setting, the variational parameters of the mixture model that best approximates the true underlying reward distribution. It attains reduced cumulative regrets when operating under complex reward models, and is valuable when restrictive modeling assumptions are undesirable. To the best of our knowledge, no other work uses variational inference to address the contextual multi-armed bandit setting. 

\clearpage
We formally introduce the contextual multi-armed bandit problem in Section \ref{sec:problem_formulation}, before providing a description of our proposed variational Thompson sampling method in Section \ref{sec:proposed_method}. We evaluate its performance in Section \ref{sec:evaluation}, and we conclude with final remarks in Section \ref{sec:conclusion}.

\section{Problem formulation}
\label{sec:problem_formulation}

The contextual multi-armed bandit problem is formulated as follows. Let $a\in\{1,\cdots,A\}$ be any possible action to take (arms in the bandit), and $f_{a}(y|x,\theta)$ the stochastic reward distribution of each arm, dependent on its intrinsic properties (\ie parameters $\theta$) and context $x\in\Real^{d}$. For every time instant $t$, the observed reward $y_t$ is independently drawn from the reward distribution corresponding to the played arm, parameterized by $\theta$ and the applicable context; \ie $y_t\sim f_{a}(y|x_t,\theta)$. We denote a set of given contexts, played arms, and observed rewards up to time instant $t$ as $x_{1:t} \equiv (x_1, \cdots , x_t)$, $a_{1:t} \equiv (a_1, \cdots , a_t)$ and $y_{1:t} \equiv (y_1, \cdots , y_t)$, respectively.

In the contextual multi-armed bandit setting, one must decide which arm to play next (\ie pick $a_{t+1}$), based on the context $x_{t+1}$, and previously observed rewards $y_{1:t}$, played arms $a_{1:t}$, and contexts $x_{1:t}$. The goal is to maximize the expected (cumulative) reward. We denote each arm's expected reward as $\mu_{a}(x,\theta)=\mathbb{E}_{a}\{y|x,\theta\}$. 

When the properties of the arms (\ie their parameters) are known, one can readily determine the optimal selection policy as soon as the context is given, \ie
\begin{equation}
a^*(x,\theta)=\argmax_{a}\mu_{a}(x,\theta) \; .
\end{equation}
The challenge in the contextual multi-armed bandit problem is raised when there is a lack of knowledge about the model. The issue amounts to the need to learn about the key properties of the environment (\ie the reward distribution), as one interacts with the world (\ie takes actions sequentially).

Amongst the many alternatives to address this class of problems, the randomized probability matching is particularly appealing. In its simplest form, known as Thompson sampling, it has been shown to perform empirically well \cite{ic-Chapelle2011, j-Scott2015} and has sound theoretical bounds, for both contextual and context-free problems \cite{ip-Agrawal2013a,ip-Agrawal2013,ip-Agrawal2012}. It plays each arm in proportion to its probability of being optimal, \ie
\begin{equation}
a_{t+1} \sim \mathrm{Pr}\left[a=a_{t+1}^*|a_{1:t}, x_{1:t+1}, y_{1:t}, \theta \right] \;.
\end{equation} 

If the parameters of the model are known, the above expression becomes deterministic, as one always picks the arm with the maximum expected reward
\begin{equation}
\mathrm{Pr}\left[a=a_{t+1}^*|a_{1:t}, x_{1:t+1}, y_{1:t},\theta \right] = \mathrm{Pr}\left[a=a_{t+1}^*|x_{t+1}, \theta \right] = I_a(x_{t+1},\theta) \;,
\label{eq:theta_known_pr_arm_optimal}
\end{equation}
where we define the indicator function $I_a(\cdot)$ as
\begin{equation}
I_a(x,\theta) = \begin{cases}
1, \; \mu_{a}(x,\theta)=\max\{\mu_1(x,\theta), \cdots, \mu_A(x,\theta)\} \;, \\
0, \; \text{otherwise} \;.
\end{cases}
\end{equation}

In practice, since the parameters of the model are unknown, one needs to explore ways of computing the probability of each arm being optimal. If the parameters are modeled as a set of random variables, then the uncertainty over the parameters can be accounted for.

Specifically, we marginalize over the posterior probability distribution of the parameters after observing rewards and actions up to time instant $t$, \ie
\begin{equation}
\begin{split}
\mathrm{Pr}\left[a=a_{t+1}^* \big| a_{1:t}, x_{1:t+1}, y_{1:t}\right] & = \int f(a|a_{1:t}, x_{1:t+1}, y_{1:t}, \theta) f(\theta|a_{1:t}, x_{1:t}, y_{1:t}) \mathrm{d}\theta \\
	& =\int I_a(x_{t+1}, \theta) f(\theta|a_{1:t}, x_{1:t}, y_{1:t}) \mathrm{d}\theta \; .
\end{split}
\label{eq:theta_unknown_pr_arm_optimal}
\end{equation}

In a Bayesian setting, if the reward distribution is known, one would assign a prior over the parameters to compute the corresponding posterior $f(\theta|a_{1:t}, x_{1:t}, y_{1:t})$. The analytical solution to such posterior is available for a well known set of distributions \cite{b-Bernardo2009}. Nevertheless, when reward distributions beyond simple well known cases (\eg Bernoulli, Gaussian, etc.) are considered, one must resort to approximations of the posterior.

In this work, we leverage variational inference to approximate such posteriors, which was founded within the discipline of statistical physics and has flourished over the past several decades in the machine learning community. 

\section{Proposed method}
\label{sec:proposed_method}

The learning process in the multi-armed bandit, as explained in the formulation of Section \ref{sec:problem_formulation}, requires updating the posterior of the reward model parameters at every time instant. For computation of $f(\theta|a_{1:t}, x_{1:t}, y_{1:t})$ in Eqn.~\eqref{eq:theta_unknown_pr_arm_optimal}, knowledge of the reward distribution is instrumental. Typically, bandit algorithms are applied to simple distributions for which sampling and calculating expectations are feasible (\eg the exponential family \cite{ic-Korda2013}).

In this work, we study finite mixture models as reward functions of the multi-armed bandit. Mixture models allow for the statistical modeling of a wide variety of stochastic phenomena; e.g., Gaussian mixture models can approximate arbitrarily well any continuous distribution and thus, provide a useful parametric framework to model unknown distributional shapes \cite{b-McLachlan2004}. This flexibility comes at a cost, as learning the parameters of the mixture distribution becomes a challenge.

We here use and empirically validate variational inference to approximate underlying Gaussian mixture models in the contextual bandit case.

For the rest of the paper, we consider a mixture of $K$ Gaussian distributions per arm $a \in \{1, \cdots, A\}$, where each of the Gaussians is linearly dependent on the shared context. Formally,
\begin{equation}
f_a(y|x,\pi_{a,k}, w_{a,k}, \sigma_{a,k}^2) =\sum_{k=1}^K \pi_{a,k} \; \N{y|x^\top w_{a,k}, \sigma_{a,k}^2}  \;,
\label{eq:gaussian_mixture_model}
\end{equation}
with per-arm mixture weights $\pi_{a,k}\in [0,1], \; \sum_{k=1}^K \pi_{a,k} = 1$ and Gaussian sufficient statistics, $w_{a,k}\in \Real^{d}$ and $\sigma_{a,k}^2\in \Real^+$.

For our analysis, we incorporate an auxiliary mixture indicator variable $z_a$. These are 1-of-K encoded vectors, where $z_{a,k}=1$, if mixture $k$ is active; $z_{a,k}=0$, otherwise.

One can now rewrite Eqn.~\eqref{eq:gaussian_mixture_model} as
\begin{equation}
f_a(y|x,z_{a},w_{a,k}, \sigma_{a,k}^2) =\prod_{k=1}^K \N{y|x^\top w_{a,k}, \sigma_{a,k}^2}^{z_{a,k}} \;,
\label{eq:gaussian_mixture_model_z}
\end{equation}
where $z_{a} \sim \Cat{\pi_{a}}$.

We consider conjugate priors for the unknown parameters of the mixture distribution
\begin{equation}
\begin{split}
f(\pi_{a}|\gamma_{a,0}) &=\Dir{\pi_a|\gamma_{a,0}} \; ,\\
f(w_{a,k}, \sigma_{a,k}^2|u_{a,k,0}, V_{a,k,0},\alpha_{a,k,0}, \beta_{a,k,0}) & = \NIG{w_{a,k}, \sigma_{a,k}^2 |u_{a,k,0}, V_{a,k,0},\alpha_{a,k,0}, \beta_{a,k,0}} \\
& = \N{w_{a,k}|u_{a,k,0}, \sigma_{a,k}^2V_{a,k,0}} \Gamma^{-1}\left(\sigma_{a,k}^2|\alpha_{a,k,0}, \beta_{a,k,0}\right). \\
\end{split}
\label{eq:mixture_model_conjugate_priors}
\end{equation}

Given a set of contexts $x_{1:t}$, played arms $a_{1:t}$, mixture assignments $z_{a,1:t}$, and observed rewards $y_{1:t}$, the joint distribution of the model follows
\begin{equation}
\begin{split}
f(y_{1:t}, z_{a,1:t}, w_{a,k}, \sigma_{a,k}^2|a_{1:t}, x_{1:t}) & = f(y_{1:t}|a_{1:t}, x_{1:t}, z_{a,1:t}, w_{a,k}, \sigma_{a,k}^2) \cdot f(z_{a,1:t}|\pi_{a}) \\
& \; \; \; \cdot f(\pi_{a}|\gamma_{a,0}) \cdot f(w_{a,k}, \sigma_{a,k}^2|u_{a,k,0}, V_{a,k,0},\alpha_{a,k,0}, \beta_{a,k,0}) \; ,
\end{split}
\end{equation}
with 
\begin{equation}
\begin{split}
f(y_{1:t}|a_{1:t}, x_{1:t}, z_{a,1:t}, w_{a,k}, \sigma_{a,k}^2)  & = \prod_{t} \prod_k \N{y_t|x_t^\top w_{a,k}, \sigma_{a,k}^2}^{z_{a,k,t}},\\
f(z_{1:t}|a_{1:t},\pi_a) &= \prod_t \prod_k \pi_{a,k}^{z_{a,k,t}},
\end{split}
\end{equation}
and parameter priors as in Eqn.~\eqref{eq:mixture_model_conjugate_priors}.

\subsection{Variational parameter inference}
\label{ssec:variational_distribution}

For the model as described above, the true joint posterior distribution is intractable. Under the variational framework, we consider instead a restricted family of distributions, and find the one that is a locally optimal approximation to the full posterior.

\begin{figure*}[!h]
	\centering
	\begin{subfigure}[b]{0.49\textwidth}
		\begin{center}
			\input{true_banditMixtureModel}
		\end{center}
		\label{fig:true_bandit}
		\caption{True contextual bandit distribution.}
	\end{subfigure}
	\begin{subfigure}[b]{0.49\textwidth}	
		\begin{center}
			\input{variational_banditMixtureModel}
		\end{center}
		\label{fig:variational_bandit}
		\caption{Variational contextual bandit distribution.}
	\end{subfigure}
	\caption{Graphical models of the bandit distribution.}
	\label{fig:graphical_bandit}
\end{figure*}
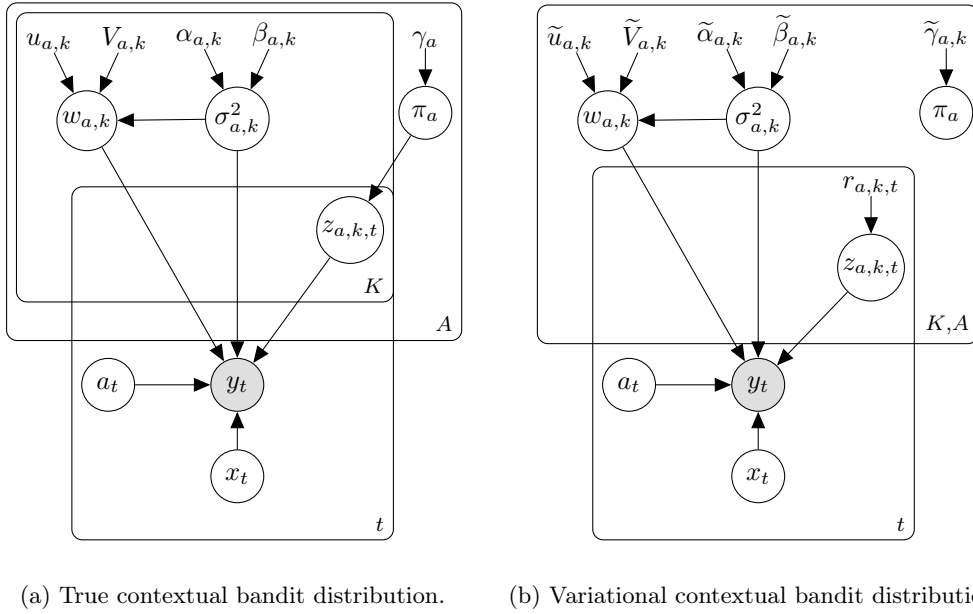

We do so by minimizing the Kullback-Leibler divergence between the true distribution $f(\cdot)$, and our approximating distribution $q(\cdot)$. We here consider a set of parameterized distributions with the following mean-field factorization over the variables of interest
\begin{equation}
q(Z, \pi, w, \sigma^2)=q(Z) \prod_{a=1}^A q(\pi_a) \prod_{k=1}^{K} q(w_{a,k}, \sigma_{a,k}^2) \; ,
\label{eq:variational_factorization}
\end{equation}
where we introduce notation $Z=\{z_{a,k,t}\}, \; \forall a,k,t$, for all latent variables; and similarly $\pi=\{\pi_{a,k}\}, \; \forall a,k$; $w=\{w_{a,k}\}, \; \forall a,k$; and $\sigma^2=\{\sigma_{a,k}^2\}, \; \forall a,k$; for parameters. We illustrate the graphical model of the true and the variational bandit distributions in Fig. \ref{fig:graphical_bandit}.

We place no restriction on the functional form of each distributional factor, and we seek to optimize the Kullback-Leibler divergence between this and the true distribution.

The optimal solution for each variational factor in the distribution in Eqn.~\eqref{eq:variational_factorization} is obtained by computing the expectation of the log-joint true distribution with respect to the rest of the variational factor distributions, as explained in \cite{b-Bishop2006}.

In our setting, we compute
\begin{equation}
\begin{split}
&\ln q(Z) =\eValue{\ln\left[f(y_{1:t}, Z, w, \sigma|a_{1:t}, x_{1:t})\right]}_{\pi, w, \sigma}+c \;, \\
&\ln q(\pi_a) =\eValue{\ln\left[f(y_{1:t}, Z, w, \sigma|a_{1:t}, x_{1:t})\right]}_{Z, w, \sigma}+c \;,\\
&\ln q(w_{a,k},\sigma_{a,k}^2) =\eValue{\ln\left[f(y_{1:t}, Z, w, \sigma|a_{1:t}, x_{1:t})\right]}_{Z,\pi}+c \;.\\
\end{split}
\end{equation}

\pagebreak
The resulting solution to the variational parameters that minimize the divergence iterates over the following two steps:
\begin{enumerate}
	\item Given the current variational parameters, compute the responsibilities
		\begin{equation}
		\begin{split}
		\log (r_{a,k,t}) &= -\frac{1}{2} \left[\ln\left(\widetilde{\beta}_{a,k}\right) - \psi \left(\widetilde{\alpha}_{a,k}\right)\right] -\frac{1}{2} \left[x_t^\top \widetilde{V}_{a,k} x_t + (y_t-x_t^\top \widetilde{u}_{a,k})^2\frac{\widetilde{\alpha}_{a,k}}{\widetilde{\beta}_{a,k}}\right] \\
		& \qquad  + \left[\psi(\widetilde{\gamma}_{a,k})- \psi\left(\sum_{k=1}^K\widetilde{\gamma}_{a,k}\right)\right] + c \;,
		\end{split}
		\end{equation}
		with $\sum_{k=1}^K r_{a,k,t} = 1$. These responsibilities correspond to the expected value of assignments, \ie $r_{a,k,t}=\eValue{z_{a,k,t}}_{Z}$.
	\item Given the current responsibilities, we define $R_{a,k}\in\Real^{t\times t}$ as a sparse diagonal matrix with diagonal elements $\left[R_{a,k}\right]_{t,t^\prime}=r_{a,k,t} \cdot \mathds{1}[a_t=a]$, and update the variational parameters
		\begin{equation}
		\begin{split}
		&\widetilde{\gamma}_{a,k}=\gamma_{a,0} + \tr{R_{a,k}} \;, \\
		&\widetilde{V}_{a,k}^{-1} = x_{1:t} R_{a,k} x_{1:t}^\top + V_{a,k,0}^{-1} \;,\\
		&\widetilde{u}_{a,k}= \widetilde{V}_{a,k} \left( x_{1:t} R_{a,k} y_{1:t} + V_{a,k,0}^{-1} u_{a,k,0}\right) \;, \\
		&\widetilde{\alpha}_{a,k} = \alpha_{a,k,0} + \frac{1}{2} \tr{R_{a,k}} \;, \\
		&\widetilde{\beta}_{a,k} = \beta_{a,k,0} + \frac{1}{2}\left(y_{1:t}^\top R_{a,k}y_{1:t} \right) + \frac{1}{2}\left( u_{a,k,0}^\top V_{a,k,0}^{-1} u_{a,k,0} - \widetilde{u}_{a,k}^\top \widetilde{V}_{a,k}^{-1} \widetilde{u}_{a,k} \right) \; .
		\end{split}
		\end{equation}
\end{enumerate}

Note that, for simplicity, we have considered the same number of mixtures per arm $K$. Nevertheless, the above expressions are readily generalizable to differing per-arm number of mixtures $K_a$, for $a\in\{1, \cdots, A\}$.

The iterative procedure presented  above is repeated until a convergence criterion is met. Usually, one iterates until the optimization improvement is small (relative to some prespecified $\epsilon$) or a maximum number of iterations is executed.

\subsection{Variational Thompson sampling}
\label{ssec:variational_thompson_sampling}

We now describe our proposed variational Thompson sampling (VTS) technique for the multi-armed contextual bandit problem, which leverages the variational distribution in subsection \ref{ssec:variational_distribution} and implements a posterior sampling based policy \cite{j-Russo2014}.

In the multi-armed bandit setting, at any given time and based on the information available, one needs to decide which arm to play next. A randomized probability matching technique picks each arm based on its probability of being optimal. In its simplest form, known as Thompson sampling \cite{j-Thompson1935}, instead of computing the integral in Eqn.~\eqref{eq:theta_unknown_pr_arm_optimal}, one draws a random parameter sample from the posterior, and then picks the action that maximizes the expected reward. That is, 
\begin{equation}
\begin{split}
a_{t+1}^*=\argmax_{a}\mu_{a}(x_{t+1},\theta_{t+1}), \qquad  \text{with} \qquad \theta_{t+1} \sim f(\theta|a_{1:t}, x_{1:t}, y_{1:t}) .
\end{split}
\end{equation}
In a pure Bayesian setting, one deals with simple models that allow for analytical computation (and sampling) of the posterior. Here, as we allow for more realistic and complex modeling of the world that may not result in closed-form posterior updates, we propose to sample the parameters from the variational approximating distributions computed in subsection \ref{ssec:variational_distribution}.

We describe the proposed variational Thompson sampling technique in Algorithm \ref{alg:vts}, for a general Gaussian mixture model with context.

\begin{algorithm}
	\begin{algorithmic}[1]
	\REQUIRE Model description $A$, $K_a$
	\REQUIRE Parameters $\gamma_{a,0}$, $u_{a,k,0}$, $V_{a,k,0}$, $\alpha_{a,k,0}$, $\beta_{a,k,0}$
	\STATE $D=\emptyset$
	\STATE Initialize $\widetilde{\gamma}_{a,k}=\gamma_{a,0}$, $\widetilde{\alpha}_{a,k}=\alpha_{a,k,0}$, $\widetilde{\beta}_{a,k}=\beta_{a,k,0}$, $ \widetilde{u}_{a,k}=u_{a,k,0}$, $\widetilde{V}_{a,k}=V_{a,k,0}$
	\FOR{$t=1, \cdots, T$}
		\STATE Receive context $x_{t+1}$
		\FOR{$a=1, \cdots, A$}
			\FOR{$k=1, \cdots, K_a$}
				\STATE Draw new parameters $\theta_{a,k,t} \deq \left\{z_{a,k,t},\pi_{a,k,t},w_{a,k,t},\sigma_{a,k,t} \right\}$
				$$\qquad \qquad \theta_{a,k,t} \sim q\left(z_{a,k},\pi_{a,k},w_{a,k},\sigma_{a,k} \left| \widetilde{\gamma}_{a,k}, \widetilde{\alpha}_{a,k}, \widetilde{\beta}_{a,k}, \widetilde{u}_{a,k}, \widetilde{V}_{a,k} \right.\right)$$
			\ENDFOR
			\STATE Compute $\mu_{a,t+1}=\mu_{a}(x_{t+1},\theta_{a,t})$
		\ENDFOR
		\STATE Play arm $a_{t+1}=\argmax_{a}\mu_{a,t+1}$
		\STATE Observe reward $y_{t+1}$
		\STATE $D=D \cup \left\{x_{t+1}, a_{t+1}, y_{t+1}\right\}$
		\WHILE{NOT Variational convergence criteria}
			\STATE Compute $r_{a,k,t}$
			\STATE Update $\widetilde{\gamma}_{a,k}, \widetilde{\alpha}_{a,k}, \widetilde{\beta}_{a,k}, \widetilde{u}_{a,k}, \widetilde{V}_{a,k}$
		\ENDWHILE
	\ENDFOR
	\end{algorithmic}
	\caption{Variational Thompson sampling}
	\label{alg:vts}
\end{algorithm}

An instrumental step in the proposed algorithm is to compute the expected reward for each arm, \ie $\mu_{a,t+1}$, for which we need both the per-arm and per-mixture parameters $\{w_{a,k}, \sigma^{2}_{a,k}\}$ and the mixture assignments $\pi_{a,k}$.
To compute the expected reward for each arm, we propose to draw per-arm and per-mixture posterior parameters from their updated variational posteriors, \ie $(w_{a,k,t}, \sigma^2_{a,k,t}) \sim q\left(w_{a,k}, \sigma^2_{a,k}|\widetilde{\alpha}_{a,k}, \widetilde{\beta}_{a,k}, \widetilde{u}_{a,k}, \widetilde{V}_{a,k}\right)$,
and consider the following mixture expectation alternatives:
\begin{enumerate}
	\item Expectation with mixture assignment sampling
	\begin{equation}
	\mu_{a,t+1}=x_{t}^\top w_{a,z_{a,k,t},t}, \; \; \;  z_{a,k,t} \sim \Cat{\frac{\widetilde{\gamma}_{a,k}}{\sum_{k=1}^{K}\widetilde{\gamma}_{a,k}} } \;.
	\end{equation}
	\item Expectation with mixture proportion sampling
	\begin{equation}
	\mu_{a,t+1}=\sum_{k=1}^K \pi_{a,k,t} x_{t}^\top w_{a,k,t}, \; \; \; \pi_{a,k,t} \sim \Dir{\widetilde{\gamma}_{a,k} } \;.
	\end{equation}
	\item Expectation with mixture proportions
	\begin{equation}
	\mu_{a,t+1}=\sum_{k=1}^K \pi_{a,k,t} x_{t}^\top w_{a,k,t}, \; \; \; \pi_{a,k,t} = \frac{\widetilde{\gamma}_{a,k}}{\sum_{k=1}^{K}\widetilde{\gamma}_{a,k}} \;.
	\end{equation}
\end{enumerate}

\section{Evaluation}
\label{sec:evaluation}

In this section, we evaluate the performance of the proposed variational Thompson sampling technique for the contextual multi-armed bandit problem.

We focus on two illustrative scenarios: the first, referred to as \texttt{Scenario A}, with per-arm reward distributions
\begin{equation}
\texttt{Scenario A }\begin{cases}
f_{0}(y|x_t,\theta) = 0.5 \cdot \N{y|(0 \; 0)^\top x_t , 1} + 0.5 \cdot \N{y|(1 \; 1)^\top x_t , 1},\\
f_{1}(y|x_t,\theta) = 0.5 \cdot \N{y|(2 \; 2)^\top x_t , 1} + 0.5 \cdot \N{y|(3 \; 3)^\top x_t , 1} \; ,
\end{cases}
\label{eq:scenario_A}
\end{equation}
and the second, \texttt{Scenario B}, with
\begin{equation}
\texttt{Scenario B }\begin{cases}
f_{0}(y|x_t,\theta) = 0.5 \cdot \N{y|(1 \; 1)^\top x_t , 1} + 0.5 \cdot \N{y|(2 \; 2)^\top x_t , 1} \; ,\\
f_{1}(y|x_t,\theta) = 0.3 \cdot \N{y|(0 \; 0)^\top x_t , 1} + 0.7 \cdot \N{y|(3 \; 3)^\top x_t , 1} \; .
\end{cases}
\label{eq:scenario_B}
\end{equation}
The reward distributions of the contextual bandits in both scenarios are Gaussian mixtures with two context dependent components.  These reward distributions are complex in that they are multimodal and, in \texttt{Scenario B}, unbalanced. Furthermore, they depend on a two dimensional uncorrelated uniform context, \ie $x_{i,t}\sim\U{0,1}$, $i\in\{1,2\}$, $t\in \Natural$.

The key difference between the scenarios is the amount of mixture overlap and the similarity between arms. Recall the complexity of the reward distributions in \texttt{Scenario B}, with a significant overlap between arm rewards and the unbalanced nature of arm 1.

We evaluate variational Thompson sampling in terms of its cumulative regret, defined as
\begin{equation}
R_t=\sum_{\tau=0}^t \eValue{\left(y^*_{\tau}-y_{\tau} \right)} = \sum_{\tau=0}^t \mu_\tau^*-\bar{y}_{\tau} \; ,
\end{equation}
where for each time instant $t$, $\mu_t^*$ denotes the true expected reward of the optimal arm, and $\bar{y}_{t}$ the empirical mean of the observed rewards.

Since we have not noticed significant cumulative regret differences between the three approaches to computing the expected reward $\mu_{a,t+1}$ described in subsection \ref{ssec:variational_thompson_sampling}, we avoid unnecessary clutter and do not plot them in the figures below. All reported values are averaged over 5000 realizations of the same set of parameters and context (with the standard deviation shown as the shaded region in the figures).

\begin{figure}[!h]
	\centering
	\begin{subfigure}[b]{0.49\textwidth}
		\includegraphics[width=\textwidth]{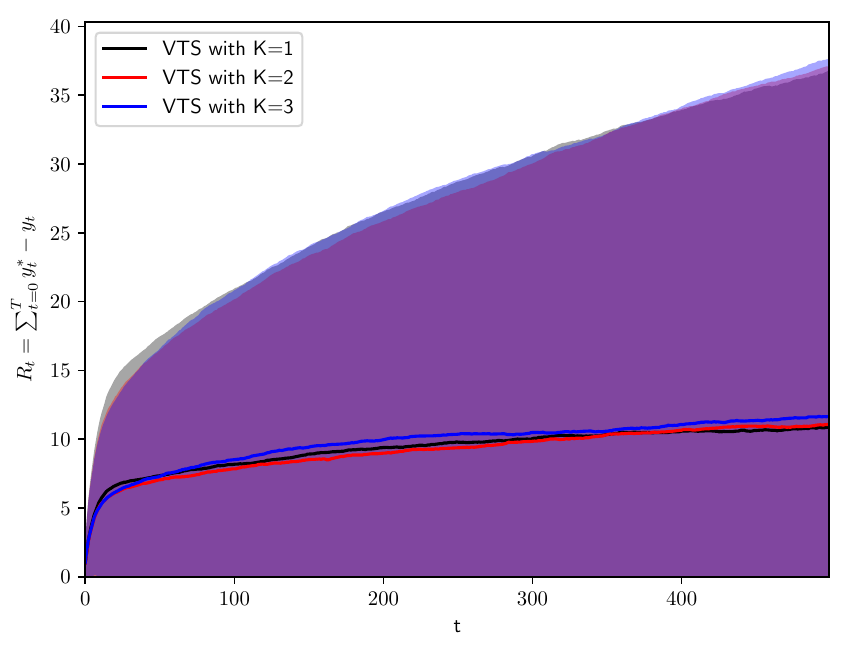}
		\caption{\texttt{Scenario A}: cumulative regret.}
		\label{fig:model_a_cumregret}
	\end{subfigure}
	\begin{subfigure}[b]{0.49\textwidth}
		\includegraphics[width=\textwidth]{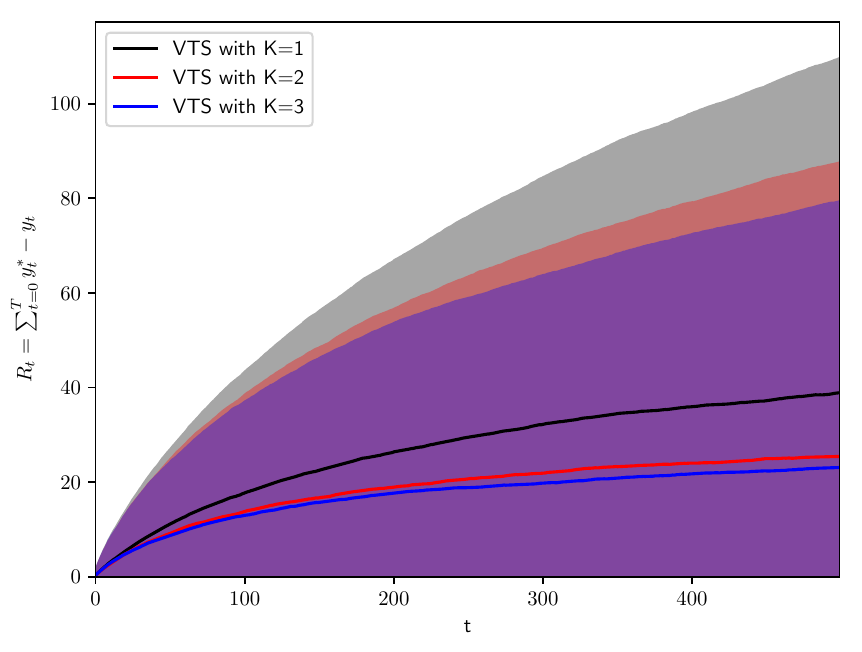}
		\caption{\texttt{Scenario B} cumulative regret.}
		\label{fig:model_b_cumregret}
	\end{subfigure}
	\caption{Cumulative regret comparison.}
	\label{fig:cumregret_comparison}
\end{figure}

Fig. \ref{fig:cumregret_comparison} shows the cumulative regret of the proposed variational Thompson sampling approach in both scenarios, when different assumptions for the variational approximating distribution are made (\ie assumed number of components $K$).

Note that ``\textit{VTS with }$K=1$'' is equivalent to a vanilla Thompson sampling approach with a linear contextual Gaussian model assumption. Since $r_{a,k=1,t}=1$ for all $a$ and $t$, the variational update equations match the corresponding Bayesian posterior updates for Thompson sampling. We are thus effectively comparing the performance of the proposed method to the Thompson sampling benchmark, as in \cite{ip-Agrawal2013a}.

The main conclusion from the results shown in Fig. \ref{fig:cumregret_comparison} is that inferring a variational approximation to the true complex reward distribution attains satisfactory regret performance.

For \texttt{Scenario A}, the regret performance of the proposed VTS with mixture of Gaussians is equivalent to ``\textit{VTS with }$K=1$'' (\ie vanilla Thompson sampling). On the contrary, for \texttt{Scenario B}, our flexible approach attains considerably lower regret. As in any posterior sampling bandit algorithm, the variance of the cumulative regret is large for all methods.

Nevertheless, we observe a reduction in both mean regret and its variability for the proposed ``\textit{VTS with }$K=2$ and $K=3$'' cases, in comparison to the contextual linear Gaussian Thompson sampling case (\ie ``\textit{VTS with }$K=1$''), for the challenging \texttt{Scenario B} illustrated in Fig. \ref{fig:model_b_cumregret}.

In other words, a misspecified (and simplified) model performs worse than the proposed (more complex) alternatives. Precisely, the cumulative regret reduction of ``\textit{VTS with }$K=2$'' (which corresponds to the true underlying mixture distributions in Eqn.~\eqref{eq:scenario_B}) with respect to ``\textit{VTS with }$K=1$'' at $t=500$ is of 35\%. The issue of model misspecification is evident for \texttt{Scenario B}, as the linear Gaussian contextual model fails to capture the subtleties of the unbalanced mixtures of Eqn.~\eqref{eq:scenario_B}.

In summary, with a simplistic model assumption as in ``\textit{VTS with }$K=1$'', one can not capture the properties of the underlying complex reward distributions and thus, can not make well-informed decisions. On the contrary, by considering more complex models (\ie Gaussian mixture models), and by using variational inference for learning its parameters, the proposed technique attains reduced regret.

Furthermore, we highlight that even an overly complex model assumption does provide competitive performance. For both \texttt{Scenario A} and \texttt{Scenario B}, the regret of the variational approximation with $K=3$ is similar to that of the true model assumption $K=2$, (``\textit{VTS with }$K=3$'' and ``\textit{VTS with }$K=2$'' in Fig. \ref{fig:cumregret_comparison}, respectively). For the challenging \texttt{Scenario B}, the cumulative regret reduction of ``\textit{VTS with }$K=3$'' with respect to the ``\textit{VTS with }$K=1$'' benchmark at $t=500$ is of 40\%. 

The explanation relies on the flexibility provided by the variational machinery, as the learning process adjusts the parameters to minimize the divergence between the true and the variational distributions. Nonetheless, one must be aware that this flexibility comes with an additional computational cost, as more parameters need to be learned.

We further elaborate on the analysis of our proposed variational Thompson sampling method by studying its learning accuracy.

In bandit algorithms, the goal is to gather enough evidence to identify the best arm (in terms of expected reward), and this can only be achieved if the arm properties (\ie the reward distributions) are learned accurately; their expectation being the most important sufficient statistic.

\begin{figure}[!h]
	\centering
	\begin{subfigure}[b]{0.49\textwidth}
		\includegraphics[width=\textwidth]{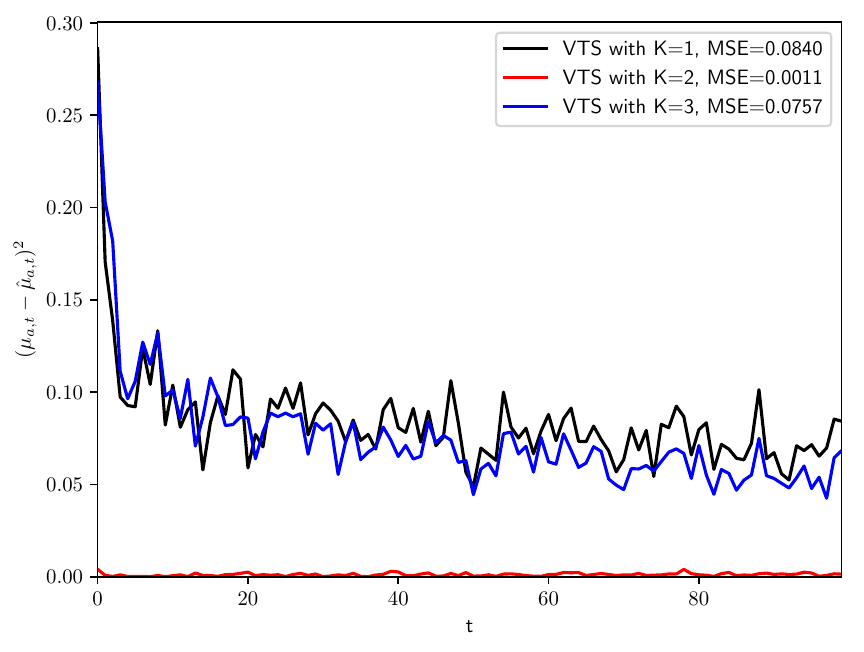}
		\caption{MSE for \texttt{Scenario A}, arm 0.}
		\label{fig:model_a_mse_arm_0}
	\end{subfigure}
	\begin{subfigure}[b]{0.49\textwidth}
		\includegraphics[width=\textwidth]{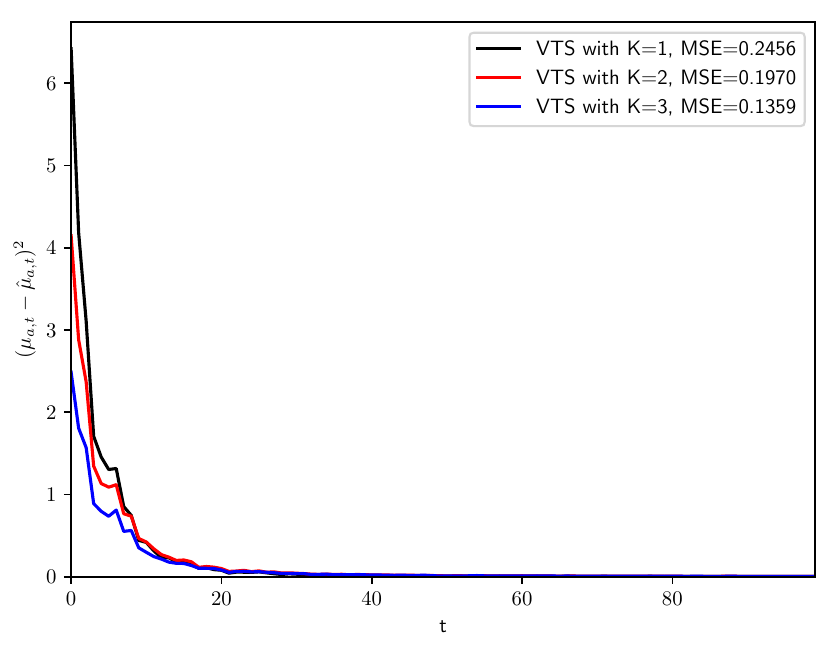}
		\caption{MSE for \texttt{Scenario A}, arm 1.}
		\label{fig:model_a_mse_arm_1}
	\end{subfigure}
	
	\begin{subfigure}[b]{0.49\textwidth}
		\includegraphics[width=\textwidth]{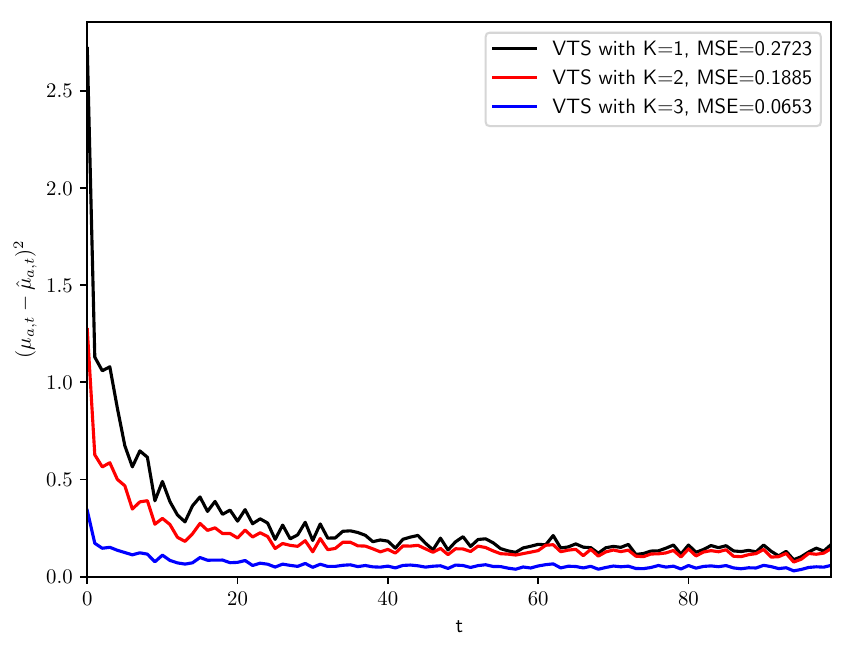}
		\caption{MSE for \texttt{Scenario B}, arm 0.}
		\label{fig:model_b_mse_arm_0}
	\end{subfigure}
	\begin{subfigure}[b]{0.49\textwidth}
		\includegraphics[width=\textwidth]{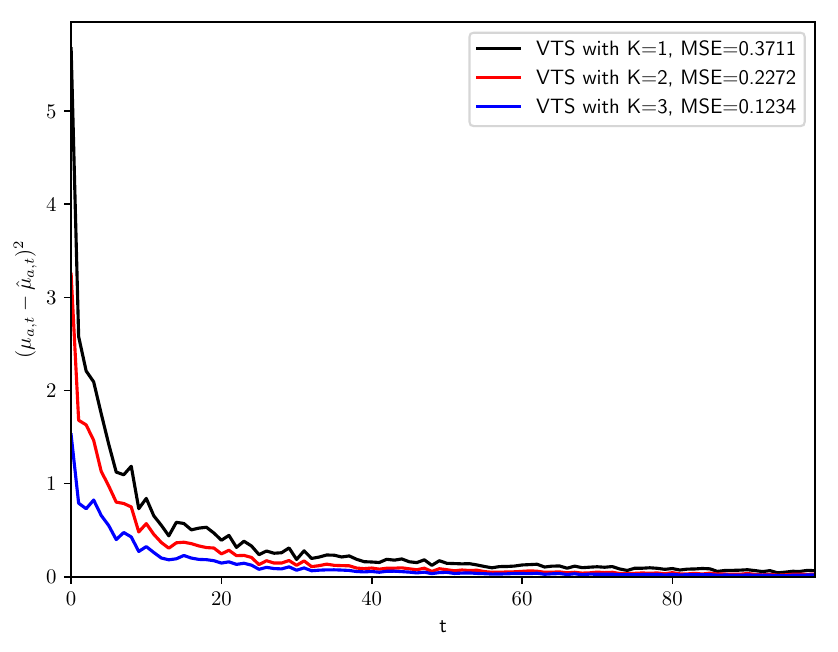}
		\caption{MSE for \texttt{Scenario B}, arm 1.}
		\label{fig:model_b_mse_arm_1}
	\end{subfigure}
	\caption{Expected reward estimation accuracy.}
	\label{fig:mse_comparison}
\end{figure}

We illustrate in Fig. \ref{fig:mse_comparison} the mean squared error of the variational per-arm expected reward estimation
\begin{equation}
MSE_a=\frac{1}{T}\sum_{t=0}^T \left(\mu_{a,t}-\hat{\mu}_{a,t} \right)^2 \; ,
\end{equation}
where $\hat{\mu}_{a,t}$ denotes the estimated expected reward for arm $a$ at time $t$.

We show that the learning is faster and more accurate when the approximating mixture model has flexibility to adapt. That is, both ``\textit{VTS with }$K=2$'' and ``\textit{VTS with }$K=3$'' accurately estimate the expected reward of the best arm.

We once again recall the complexity of the model in \texttt{Scenario B} in comparison to that of \texttt{Scenario A}, and more importantly, its implications for a bandit algorithm. In Figs. \ref{fig:model_a_mse_arm_0}-\ref{fig:model_a_mse_arm_1}, the simplest model that assumes a single Gaussian distribution (``\textit{VTS with }$K=1$'') is able to quickly and accurately estimate the expected reward. In contrast, its estimation accuracy is the worst (as shown in Figs. \ref{fig:model_b_mse_arm_0}-\ref{fig:model_b_mse_arm_1}) when facing a more complex model with overlapping and unbalanced arm rewards. Note how, for all results in Fig. \ref{fig:mse_comparison}, the most complex model (\ie ``\textit{VTS with }$K=3$'') fits the expected reward best.

These observations reinforce our claims on the flexibility and applicability of the presented technique. By allowing for complex modeling of the world and using variational inference to learn it, the proposed variational Thompson sampling technique can provide improved performance (in the sense of regret) for the contextual multi-armed bandit problem.

\section{Conclusion}
\label{sec:conclusion}

We have presented variational Thompson sampling, a new algorithm for the contextual multi-armed bandit setting, where we combine variational inference machinery with a state of the art reinforcement learning technique. The proposed algorithm allows for interpretable bandit modeling with complex reward functions, learned from online data. We extend the applicability of Thompson sampling by accommodating more realistic and complex models of the world. Empirical results show a significant cumulative regret reduction when using the proposed algorithm in simulated models. A natural future application is to scenarios when relevant context (attributes of items, customers or patients) are unobservable, and thus the latent variables are truly `incomplete' as in the motivating case for expectation maximization modeling \cite{j-Dempster1977}.

\subsection{Software and Data}
The implementation of the proposed method is available in \href{https://github.com/iurteaga/bandits}{this public repository}. It contains all the software required for replication of the findings of this study.

\subsubsection*{Acknowledgments}
This research was supported in part by NSF grant SCH-1344668. We thank Shipra Agrawal, David Blei and Daniel J. Hsu for discussions that helped motivate this work. We also thank Hang Su, Edward Peng Yu, and all the reviewers for their feedback and comments.


\input{vts.bbl}
\bibliographystyle{abbrvnat}

\end{document}

%% file: true_banditMixtureModel.tex
\begin{tikzpicture}
	\node[obs] (y-t) {$y_{t}$};
	\node[latent, above=2.75 of y-t, xshift=-2cm]  (w-ak) {$w_{a,k}$};
	\node[latent, above=2.75 of y-t, xshift=0cm] (sigma-ak) {$\sigma_{a,k}^2$};
	\node[latent, above=1.25 of y-t, xshift=1.5cm] (z-akt) {$z_{a,k,t}$};
	\node[latent, above=0.75 of z-akt, xshift=1cm] (pi-a) {$\pi_{a}$};	
	\node[latent, left=1 of y-t] (a-t) {$a_t$};
	\node[latent, below=0.5 of y-t]  (x-t) {$x_t$};
	
	\node[const, above=0.5 of sigma-ak, xshift=-0.5cm] (alpha-ak) {$\alpha_{a,k}$} ;
	\node[const, above=0.5 of sigma-ak, xshift=0.5cm]  (beta-ak) {$\beta_{a,k}$} ;
	
	\node[const, above=0.5 of w-ak, xshift=-0.5cm] (mu-ak) {$u_{a,k}$} ;
	\node[const, above=0.5 of w-ak, xshift=0.5cm]  (V-ak) {$V_{a,k}$} ;
	
	\node[const, above=0.5 of pi-a] (gamma-a) {$\gamma_{a}$} ;
	
	\edge {alpha-ak,beta-ak} {sigma-ak} ;
	\edge {mu-ak,V-ak,sigma-ak} {w-ak} ;
	\edge {gamma-a} {pi-a} ;
	\edge {pi-a} {z-akt} ;
	\edge {sigma-ak, w-ak, z-akt,x-t,a-t} {y-t} ;
	
	\plate {t} {(a-t)(x-t)(z-akt)(y-t)} {$t$} ;
	\plate {k}{
		(alpha-ak)(beta-ak)(mu-ak)(V-ak) 
		(sigma-ak)(w-ak) 
		(z-akt) 
		} {$K$} ;
	\plate {a}{
		(alpha-ak)(beta-ak)(mu-ak)(V-ak)(gamma-a) 
		(sigma-ak)(w-ak)(pi-a) 
		(z-akt) 
		(k.north west) (k.south west) 
	} {$A$} ;
\end{tikzpicture}

%% file: variational_banditMixtureModel.tex
\begin{tikzpicture}
	\node[obs] (y-t) {$y_{t}$};
	\node[latent, above=2.75 of y-t, xshift=-2cm]  (w-ak) {$w_{a,k}$};
	\node[latent, above=2.75 of y-t, xshift=0cm] (sigma-ak) {$\sigma_{a,k}^2$};
	\node[latent, above=0.75 of y-t, xshift=1.5cm] (z-akt) {$z_{a,k,t}$};
	\node[latent, above=1.25 of z-akt, xshift=1cm] (pi-a) {$\pi_{a}$};	
	\node[latent, left=1 of y-t] (a-t) {$a_t$};
	\node[latent, below=0.5 of y-t]  (x-t) {$x_t$};
	
	\node[const, above=0.5 of sigma-ak, xshift=-0.5cm] (tildealpha-ak) {$\widetilde{\alpha}_{a,k}$} ;
	\node[const, above=0.5 of sigma-ak, xshift=0.5cm]  (tildebeta-ak) {$\widetilde{\beta}_{a,k}$} ;
	
	\node[const, above=0.5 of w-ak, xshift=-0.5cm] (tildew-ak) {$\widetilde{u}_{a,k}$} ;
	\node[const, above=0.5 of w-ak, xshift=0.5cm]  (tildeV-ak) {$\widetilde{V}_{a,k}$} ;
	
	\node[const, above=0.5 of pi-a] (tildegamma-ak) {$\widetilde{\gamma}_{a,k}$} ;

	\node[const, above=0.5 of z-akt] (r-akt) {$r_{a,k,t}$} ;
	
	\edge {tildealpha-ak,tildebeta-ak} {sigma-ak} ;
	\edge {tildew-ak,tildeV-ak,sigma-ak} {w-ak} ;
	\edge {tildegamma-ak} {pi-a} ;
	\edge {sigma-ak, w-ak, z-akt,x-t,a-t} {y-t} ;
	\edge {r-akt} {z-akt} ;
	
	\plate {t} {(a-t)(x-t)(z-akt)(y-t)(r-akt)} {$t$} ;
	\plate {k}{
		(tildealpha-ak)(tildebeta-ak)(tildew-ak)(tildeV-ak)(tildegamma-ak) 
		(sigma-ak)(w-ak) 
		(z-akt)(r-akt) 
		} {$K$,$A$} ;
\end{tikzpicture}

%% file: vts.bbl
\begin{thebibliography}{23}
\providecommand{\natexlab}[1]{#1}
\providecommand{\url}[1]{\texttt{#1}}
\expandafter\ifx\csname urlstyle\endcsname\relax
  \providecommand{\doi}[1]{doi: #1}\else
  \providecommand{\doi}{doi: \begingroup \urlstyle{rm}\Url}\fi

\bibitem[Agrawal and Goyal(2012)]{ip-Agrawal2012}
S.~Agrawal and N.~Goyal.
\newblock {Analysis of Thompson Sampling for the multi-armed bandit problem}.
\newblock In \emph{{Conference on Learning Theory}}, pages 39--1, 2012.

\bibitem[Agrawal and Goyal(2013{\natexlab{a}})]{ip-Agrawal2013}
S.~Agrawal and N.~Goyal.
\newblock {Further Optimal Regret Bounds for Thompson Sampling}.
\newblock In \emph{{Artificial Intelligence and Statistics}}, pages 99--107,
  2013{\natexlab{a}}.

\bibitem[Agrawal and Goyal(2013{\natexlab{b}})]{ip-Agrawal2013a}
S.~Agrawal and N.~Goyal.
\newblock {Thompson Sampling for Contextual Bandits with Linear Payoffs}.
\newblock In \emph{{International Conference on Machine Learning}}, pages
  127--135, 2013{\natexlab{b}}.

\bibitem[Bernardo and Smith(2009)]{b-Bernardo2009}
J.~M. Bernardo and A.~F. Smith.
\newblock \emph{{Bayesian Theory}}.
\newblock Wiley Series in Probability and Statistics. Wiley, 2009.
\newblock ISBN 9780470317716.
\newblock \doi{10.1002/9780470316870}.

\bibitem[Bishop(2006)]{b-Bishop2006}
C.~Bishop.
\newblock \emph{{Pattern Recognition and Machine Learning}}.
\newblock Information Science and Statistics. Springer-Verlag New York, 2006.

\bibitem[Blundell et~al.(2015)Blundell, Cornebise, Kavukcuoglu, and
  Wierstra]{ip-Blundell2015}
C.~Blundell, J.~Cornebise, K.~Kavukcuoglu, and D.~Wierstra.
\newblock {Weight Uncertainty in Neural Networks}.
\newblock In \emph{Proceedings of the 32Nd International Conference on
  International Conference on Machine Learning - Volume 37}, ICML'15, pages
  1613--1622. JMLR.org, 2015.

\bibitem[Chapelle and Li(2011)]{ic-Chapelle2011}
O.~Chapelle and L.~Li.
\newblock {An Empirical Evaluation of Thompson Sampling}.
\newblock In J.~Shawe-Taylor, R.~S. Zemel, P.~L. Bartlett, F.~Pereira, and
  K.~Q. Weinberger, editors, \emph{Advances in Neural Information Processing
  Systems 24}, pages 2249--2257. Curran Associates, Inc., 2011.
\newblock URL
  \url{https://papers.nips.cc/paper/4321-an-empirical-evaluation-of-thompson-sampling}.

\bibitem[Dempster et~al.(1977)Dempster, Laird, and Rubin]{j-Dempster1977}
A.~P. Dempster, N.~M. Laird, and D.~B. Rubin.
\newblock {Maximum likelihood from incomplete data via the EM algorithm}.
\newblock \emph{Journal of the royal statistical society. Series B
  (methodological)}, pages 1--38, 1977.

\bibitem[Ghavamzadeh et~al.(2015)Ghavamzadeh, Mannor, Pineau, and
  Tamar]{j-Ghavamzadeh2015}
M.~Ghavamzadeh, S.~Mannor, J.~Pineau, and A.~Tamar.
\newblock {Bayesian Reinforcement Learning: A Survey}.
\newblock \emph{Foundations and Trends® in Machine Learning}, 8\penalty0
  (5-6):\penalty0 359--483, 2015.
\newblock ISSN 1935-8237.
\newblock \doi{10.1561/2200000049}.

\bibitem[Korda et~al.(2013)Korda, Kaufmann, and Munos]{ic-Korda2013}
N.~Korda, E.~Kaufmann, and R.~Munos.
\newblock {Thompson Sampling for 1-Dimensional Exponential Family Bandits}.
\newblock In C.~J.~C. Burges, L.~Bottou, M.~Welling, Z.~Ghahramani, and K.~Q.
  Weinberger, editors, \emph{Advances in Neural Information Processing Systems
  26}, pages 1448--1456. Curran Associates, Inc., 2013.

\bibitem[Li et~al.(2010)Li, Chu, Langford, and Schapire]{ip-Li2010}
L.~Li, W.~Chu, J.~Langford, and R.~E. Schapire.
\newblock {A Contextual-Bandit Approach to Personalized News Article
  Recommendation}.
\newblock In \emph{Proceedings of the 19th international conference on World
  wide web}, volume abs/1003.0146, pages 661--670, 2010.

\bibitem[Lipton et~al.(2018)Lipton, Li, Gao, Li, Ahmed, and
  Deng]{ip-Lipton2018}
Z.~C. Lipton, X.~Li, J.~Gao, L.~Li, F.~Ahmed, and L.~Deng.
\newblock {BBQ-Networks: Efficient Exploration in Deep Reinforcement Learning
  for Task-Oriented Dialogue Systems}.
\newblock In \emph{AAAI}, 2018.

\bibitem[Maillard et~al.(2011)Maillard, Munos, and Stoltz]{ip-Maillard2011}
O.-A. Maillard, R.~Munos, and G.~Stoltz.
\newblock {Finite-Time Analysis of Multi-armed Bandits Problems with
  Kullback-Leibler Divergences}.
\newblock In \emph{Conference On Learning Theory}, 2011.

\bibitem[McLachlan and Peel(2004)]{b-McLachlan2004}
G.~McLachlan and D.~Peel.
\newblock \emph{{Finite Mixture Models}}.
\newblock Wiley Series in Probability and Statistics. John Wiley \& Sons, 2004,
  2004.
\newblock ISBN 9780471654063.

\bibitem[Osband et~al.(2016)Osband, Blundell, Pritzel, and Roy]{ic-Osband2016}
I.~Osband, C.~Blundell, A.~Pritzel, and B.~V. Roy.
\newblock {Deep Exploration via Bootstrapped DQN}.
\newblock In D.~D. Lee, M.~Sugiyama, U.~V. Luxburg, I.~Guyon, and R.~Garnett,
  editors, \emph{Advances in Neural Information Processing Systems 29}, pages
  4026--4034. Curran Associates, Inc., 2016.

\bibitem[Robbins(1952)]{j-Robbins1952}
H.~Robbins.
\newblock {Some aspects of the sequential design of experiments}.
\newblock \emph{Bulletin of the American Mathematical Society}, \penalty0
  (58):\penalty0 527--535, 1952.

\bibitem[Russo and Roy(2014)]{j-Russo2014}
D.~Russo and B.~V. Roy.
\newblock {Learning to optimize via posterior sampling}.
\newblock \emph{Mathematics of Operations Research}, 39\penalty0 (4):\penalty0
  1221--1243, 2014.

\bibitem[Russo and Roy(2016)]{j-Russo2016}
D.~Russo and B.~V. Roy.
\newblock {An information-theoretic analysis of Thompson sampling}.
\newblock \emph{The Journal of Machine Learning Research}, 17\penalty0
  (1):\penalty0 2442--2471, 2016.

\bibitem[Scott(2010)]{j-Scott2010}
S.~L. Scott.
\newblock {A modern Bayesian look at the multi-armed bandit}.
\newblock \emph{Applied Stochastic Models in Business and Industry},
  26\penalty0 (6):\penalty0 639--658, 2010.
\newblock ISSN 1526-4025.
\newblock \doi{10.1002/asmb.874}.

\bibitem[Scott(2015)]{j-Scott2015}
S.~L. Scott.
\newblock {Multi-armed bandit experiments in the online service economy}.
\newblock \emph{Applied Stochastic Models in Business and Industry},
  31:\penalty0 37--49, 2015.
\newblock Special issue on actual impact and future perspectives on stochastic
  modelling in business and industry.

\bibitem[Sutton and Barto(1998)]{b-Sutton1998}
R.~S. Sutton and A.~G. Barto.
\newblock \emph{{Reinforcement Learning: An Introduction}}.
\newblock MIT Press: Cambridge, MA, 1998.

\bibitem[Thompson(1933)]{j-Thompson1933}
W.~R. Thompson.
\newblock {On the Likelihood that One Unknown Probability Exceeds Another in
  View of the Evidence of Two Samples}.
\newblock \emph{Biometrika}, 25\penalty0 (3/4):\penalty0 285--294, 1933.
\newblock ISSN 00063444.

\bibitem[Thompson(1935)]{j-Thompson1935}
W.~R. Thompson.
\newblock {On the Theory of Apportionment}.
\newblock \emph{American Journal of Mathematics}, 57\penalty0 (2):\penalty0
  450--456, 1935.
\newblock ISSN 00029327, 10806377.

\end{thebibliography}
